%% file: 00_main.tex
\documentclass[preprint,journal]{vgtc}                  

\preprinttext{To appear in IEEE Transactions on Visualization and Computer Graphics.}

\vgtccategory{Research}

\vgtcpapertype{System}

\title{Visualizing and Interpreting Temporal Care Team Dynamics for Cancer Survival Risk Analysis}

\author{
Yuhua Huang\orcidlink{0009-0001-0477-8341},
Hsiao-Ying Lu\orcidlink{0009-0000-1145-9229},
and Kwan-Liu Ma \orcidlink{0000-0001-8086-0366} \textit{Fellow, IEEE}
}
\authorfooter{
\item
Yuhua Huang, Hsiao-Ying Lu, and Kwan-Liu Ma are with the Department of Computer Science, University of California, Davis.
E-mail: \{yuuhuang, hyllu, klma\}@ucdavis.edu.
}

\input{0_abstract}


\graphicspath{{figs/}{figures/}{pictures/}{images/}{./}} 

\usepackage{tabu}                      
\usepackage{booktabs}                  
\usepackage{lipsum}                    
\usepackage{mwe}                       

\usepackage{mathptmx}                  
\usepackage{cite}
\usepackage{amsmath,amsfonts}
\usepackage{algorithmic}
\usepackage{array}
\usepackage[caption=false,font=normalsize,labelfont=sf,textfont=sf]{subfig}
\usepackage{textcomp}
\usepackage{stfloats}
\usepackage{url}
\usepackage{verbatim}
\usepackage{graphicx}
\usepackage{balance}

\usepackage{xcolor}
\usepackage{soul}

\soulregister\cite7
\soulregister\autoref7
\soulregister\ref7
\soulregister\pageref7

\soulregister\parencite7
\soulregister\textcite7

\usepackage{xcolor}
\usepackage[pagebackref,bookmarks]{hyperref}
\usepackage{tikz}
\usetikzlibrary{arrows.meta,positioning,fit,calc,backgrounds,shapes.geometric}

\usepackage{orcidlink}
\begin{document}


\firstsection{Introduction}

\maketitle

\input{1_introduction}
\input{2_related}

\input{3_methodology}

\input{4_case_study}

\input{5_expert_feedback}
\input{6_discussion}
\input{7_conclusion}
\acknowledgments{
     This research was supported by the National Institutes of Health through National Cancer Institute grants R01 CA270454 and R01 CA273058. The content is solely the responsibility of the authors and does not necessarily represent the official views of the National Institutes of Health.

     We thank Shin-Ping Tu, MD, MPH and Brittany Garcia, PhD (UC Davis); Michael Hogarth, MD and Aaron Boussina, PhD (UC San Diego); Daniel K.  Sewell, PhD (University of Iowa); and Khalid Matin, MD (Virginia
    Commonwealth University), together with the wider SMART Cancer Care
    Teams investigators, for the clinical, informatics, and methodological
    guidance that shaped this work.

}

\bibliographystyle{abbrv-doi-hyperref}
\bibliography{00_ref}

\end{document}

%% file: 0_abstract.tex
\abstract{
Cancer care is a longitudinal, time-sensitive process that depends on the coordinated activities of diverse healthcare professionals (HCPs), whose patterns of collaboration evolve continuously throughout the course of treatment. Prior studies have shown that care team structure is associated with patient outcomes; however, the temporal dynamics of these collaborations and their relationship to evolving patient risk remain insufficiently understood. To address this gap, we introduce a visual analytics framework for modeling and interpreting time-varying care team interactions in oncology. We construct dynamic collaboration networks from electronic health record (EHR)-mediated communication data and employ a temporal graph neural network (TGNN) to estimate patient outcome risk over time. Building on these predictions, we examine how changes in care team composition and specialist participation are associated with shifts in estimated risk across different treatment stages. To support practitioner-driven analysis, we design a multi-view interactive system that integrates TGNN-based risk modeling with coordinated visualizations of care team networks, feature-level explanations derived from a surrogate logistic regression model, and exploratory what-if analysis. Rather than inferring causality, the system enables analysts to investigate how risk estimates co-evolve with collaboration patterns and to identify the care team characteristics most strongly associated with these temporal changes. We demonstrate the utility of the system through expert-guided analysis scenarios on a cohort of 505 cancer patients spanning breast, colorectal, and lung cancers, and evaluate its effectiveness in uncovering clinically meaningful collaboration patterns. This work contributes a practical approach for improving the interpretability of temporal graph-based models in longitudinal, team-based healthcare settings, with broader applicability to domains involving complex, evolving collaborative processes.
}

%% file: 1_introduction.tex
Understanding how clinical teams collaborate over time is critical to improving patient care, particularly in complex domains such as cancer treatment where patients interact with multiple healthcare professionals (HCPs) across extended care trajectories \cite{fleissig2006multidisciplinary, gurses2006systematic, Smits2010-ok, bagnasco2013identifying, verhaegh2017exploratory}. Prior work has examined the relationship between care team structures and patient outcomes, often modeling collaboration as static networks aggregated over an entire episode of care \cite{barnett2011mapping, Zhu2019-tu}. However, real-world care delivery is inherently dynamic: team composition, roles, and interactions evolve as patient conditions change, motivating the need for representations that capture these temporal dynamics. Electronic health records (EHRs) provide a promising foundation for this purpose. As the primary platform for documenting and coordinating care, EHRs generate fine-grained, timestamped logs that reflect healthcare professionals’ activity around patient care. While most prior research has focused on patient-level physiological data within EHRs \cite{huang2017regularized, amirahmadi2023deep, nelson2019integrating}, relatively little attention has been given to the metadata generated through system usage---such as access logs, shared documentation, or co-signatures---as signals of clinical collaboration. These underutilized digital traces offer a scalable and data-driven means to construct time-evolving collaboration networks, providing new opportunities to study how care teams function in real-world settings and how their dynamics relate to patient outcomes.




In this work, we leverage timestamped EHR logs to extract digital traces of EHR-mediated collaboration among HCPs, constructing patient-specific, time-evolving collaboration networks that represent how care teams form and evolve throughout treatment~\cite{zhu2022ehr, yao2018ehr}. Reasoning about these dynamics requires bringing together three signals that are often examined separately: how collaboration structure changes over time, how model-estimated patient risk changes alongside it, and which care-team characteristics are highlighted by model explanations. Prior work has shown that care-team structure is associated with patient outcomes \cite{barnett2011mapping, Zhu2019-tu}, and recent graph-based approaches further demonstrate that EHR-derived collaboration patterns can support outcome prediction \cite{huang2026dynamics, lu2025associatinghealthcareteamworkpatient}. 

However, predictive modeling, visual exploration of collaboration, and feature-level interpretation are still often treated as separate steps or represented using different temporal bases. Because care delivery is itself dynamic, with team composition, roles, and interactions evolving as patient conditions change \cite{bagnasco2013identifying, fleissig2006multidisciplinary,gurses2006systematic}, risk estimates and explanations are more interpretable when they are aligned with the same evolving collaboration structures shown to analysts. To address this gap, we model patient risk using a temporal graph neural network \cite{longa2023graph} over the same time-evolving collaboration networks used in the visual analysis, providing a shared temporal basis for risk estimation, explanation, and interactive exploration.


To build on this alignment between temporal collaboration modeling and risk interpretation, we introduce a visual analytics system that couples model-estimated risk trajectories with coordinated, interactive views of evolving care-team networks. The system supports practitioners in moving between cohort-level comparison, patient-level temporal inspection, and feature-level interpretation, enabling them to examine how collaboration patterns and model-estimated risk co-evolve over the treatment timeline. We describe the individual views, interactions, and analytical tasks in \autoref{sec:frontend}.
Importantly, our goal is not to establish definitive clinical conclusions, but to support insight generation and hypothesis development. Through case studies and expert feedback, we demonstrate how the system enables clinicians to explore complex relationships between collaboration dynamics and patient risk trajectories, uncover patterns of interest, and generate hypotheses for further investigation. In doing so, we position our work as a step toward more transparent and interpretable integration of predictive modeling into clinical sensemaking workflows.


In summary, this work makes the following contributions:
\begin{itemize}[nosep, leftmargin=1.8em]
\item A characterization of the analytical tasks for interpreting temporally evolving care-team collaboration in relation to model-estimated patient risk, developed with clinical and informatics collaborators.
\item A visual analytics framework that aligns temporal graph-based risk modeling, interpretable feature attribution using a surrogate model, and coordinated interactive exploration on a shared temporal basis, addressing the mismatch in which dynamic collaboration views are paired with risk and explanations derived from static or aggregated inputs.
\item A coordinated multi-view system that instantiates these tasks---supporting cohort-level, comparative, and longitudinal analysis together with exploratory what-if analysis---evaluated through expert review and case studies that demonstrate practitioner-driven interpretation, insight generation, and hypothesis development.
\end{itemize}

%% file: 2_related.tex
\section{Related Work}
\label{sec:relatedworks}

\subsection{Healthcare Team Collaboration Analysis}

Understanding team-based care in healthcare has long been an important topic in healthcare services research. Prior work has shown that team composition, coordination, and information exchange can affect care delivery and patient outcomes, particularly in complex clinical settings where patients interact with multiple providers across extended care trajectories \cite{fleissig2006multidisciplinary, Smits2010-ok, gurses2006systematic, bagnasco2013identifying, verhaegh2017exploratory}. Related work on patient-sharing and physician networks further demonstrated that structured clinical records can be used to recover meaningful professional relationships among healthcare providers \cite{barnett2011mapping}.

With the widespread adoption of electronic health record (EHR) systems, researchers have increasingly used digital traces of clinical activity to study care-team collaboration at finer granularity. Prior work has shown that EHR access-log data can be used to construct patient-centered and virtual care-team networks, revealing heterogeneity in care-team structure and communication patterns across patients and clinical contexts \cite{yao2018ehr, zhu2022ehr, Zhu2019-tu}. Visual analytics systems have also been introduced to support exploratory analysis of EHR-mediated communication and collaboration networks \cite{10614409}.

Together, these studies established that EHR-derived care-team networks provide a useful representation of EHR-mediated collaboration and support exploratory analysis of team structure. However, most prior work has emphasized descriptive social-network analysis or aggregate communication measures, leaving limited support for interactive interpretation of patient-level temporal dynamics in predictive modeling settings.

\subsection{Modeling and Interpreting Dynamic Graphs}

\subsubsection{Modeling dynamic graphs}

Dynamic graph learning has been widely studied in the machine learning community through embedding-based, recurrent, and attention-based methods. Early work such as DynGEM \cite{goyal2018dyngem} modeled time-evolving node representations, while later methods such as DySAT \cite{sankar2020dysat} incorporated temporal modeling directly into graph neural architectures, and EvolveGCN \cite{pareja2020evolvegcn} evolves graph convolution parameters over time.

Within healthcare, recent work has begun applying graph-based learning to EHR-derived collaboration networks for patient outcome prediction \cite{lu2025associatinghealthcareteamworkpatient, huang2026dynamics}. Some approaches model collaboration within a fixed observation window, emphasizing predictive signals embedded in aggregated structure and node attributes, whereas more recent temporal formulations represent collaboration as a sequence of cumulative weekly snapshots to capture how care teams expand and change over the course of treatment \cite{huang2026dynamics}. These studies show that evolving collaboration structure can be modeled computationally, but they leave open the interpretability challenge of understanding how changing collaboration patterns relate to evolving patient risk and model behavior over time.

\subsubsection{Interpreting dynamic graphs}
As graph neural networks have become more widely adopted, a substantial body of literature has emerged on explainability for graph-based models \cite{ying2019gnnexplainer, luo2020pgexplainer, yuan2021subgraphx, xia2023tgnexplainer, yuan2023gnnsurvey, lu2024gnnanatomy}. Among different approaches, surrogate-based methods use simpler, interpretable models to support explanations in terms of human-understandable features. GraphLIME, for example, uses a local surrogate model to explain graph predictions through interpretable features \cite{huang2020graphlime}. In healthcare settings, where analysts reason in terms of provider roles, specialties, and documentation behaviors, such feature-centered explanations are particularly useful, as they align with domain-relevant concepts and are more actionable than explanations based solely on subgraph structures.

Yet, feature-centered explanations alone are often insufficient for supporting temporal analytical tasks. While they can reveal which interpretable features are associated with a prediction at a given time point, they typically do not support longitudinal comparison, temporal drill-down, or reasoning about how feature contributions evolve alongside changes in collaboration structure. This limitation highlights the need for approaches that situate feature-based explanations within a broader temporal and structural analysis context, enabling users to relate model explanations to the underlying dynamics of the data.

\subsubsection{Visual analytics systems for graphs}
Visual analytics provides a powerful approach for exploring complex relationships and generating evidence to support user-driven reasoning. Such interactive systems enable users to interpret graph neural networks, complex predictive models, and time-varying data through coordinated views and interactive exploration. For example, GNNLens and CorGIE support the inspection of GNN predictions, embeddings, and graph structures \cite{jin2020gnnlens, liu2022corgie}. In healthcare, RetainVis demonstrates how visual analytics can facilitate the interpretation of temporal predictive models over electronic medical records \cite{kwon2019retainvis}. Within the care-team collaboration domain, EHRFlow enables the exploration of communication effectiveness and efficiency in EHR-mediated collaboration networks \cite{10614409}.  

Recent healthcare visual analytics systems have demonstrated the value of interactive visualization for helping domain experts interpret longitudinal clinical data, disease trajectories, and EHR-derived activity patterns. DPVis integrates Hidden Markov Models with visual summaries to support exploration of disease progression states, pathways, and patient subgroups \cite{9058722}. ClinicalPath supports longitudinal EHR interpretation by helping clinicians review patient histories, laboratory results, and clinical events for decision-making \cite{9779066}. Prior work on healthcare team communication networks further shows that EHR-derived interaction traces can be transformed into visual network representations for analyzing care-team communication patterns \cite{lu2023icmhi}. 
These systems establish the value of interactive visual reasoning for clinical-trajectory and healthcare-collaboration analysis, but they do not jointly support temporal collaboration networks, model-estimated risk, and feature-level explanations on a shared temporal basis. This limits practitioners' ability to reason about how evolving care-team structures relate to changes in predicted risk and the feature attributions associated with those predictions.

%% file: 3_methodology.tex
        
\section{Methodology}
\label{sec:methodology}

\begin{figure*}[t]
    \centering
    \includegraphics[width=\textwidth]{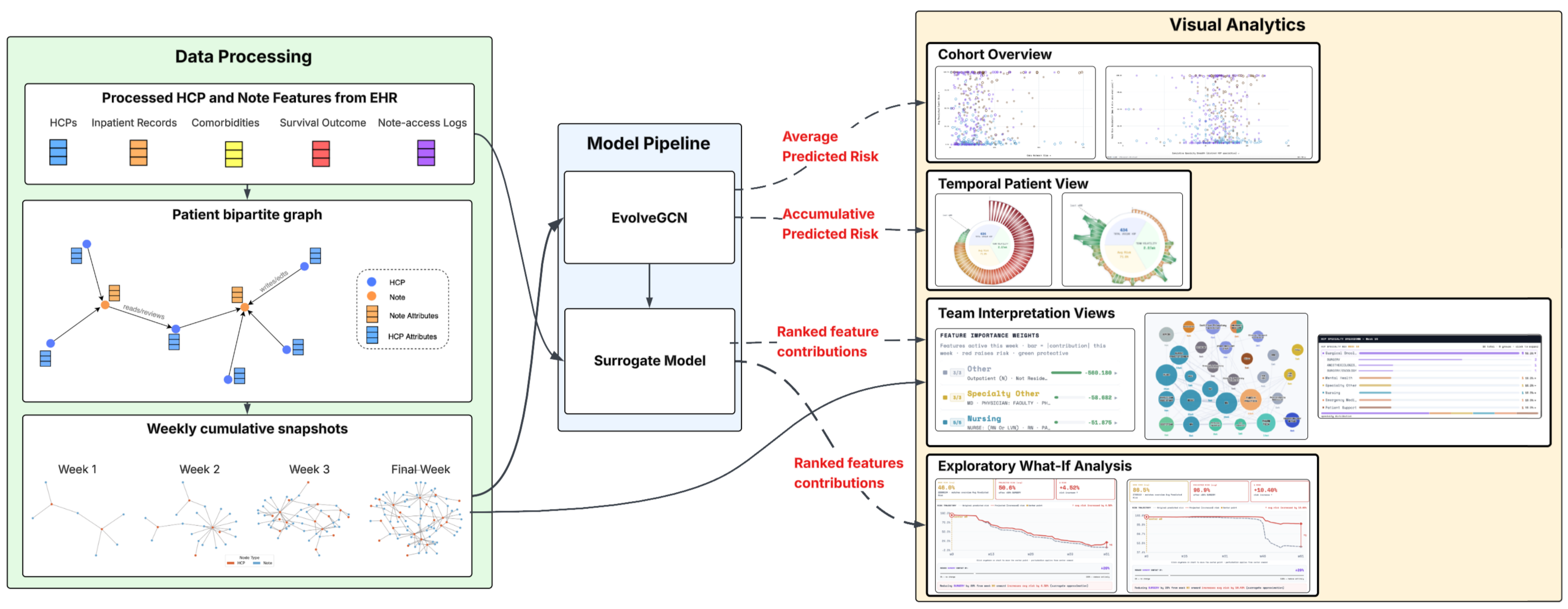}
    \caption{Overview of the methodology pipeline, showing how curated EHR data are transformed into temporal risk estimates, surrogate-based feature attributions, and coordinated visual representations. The arrangement summarizes the system components and their outputs rather than the analyst-facing workflow.}
    \label{fig:flowchart}
    \vspace{-10pt}
\end{figure*}

To analyze time-evolving clinical collaboration and model-estimated risk, we integrate temporal collaboration networks, risk estimation, surrogate-based feature attribution, and coordinated visual analysis. \autoref{fig:flowchart} summarizes these system components, while the analyst-facing workflow is described in \autoref{sec:frontend}.

\subsection{Data Processing}
\label{sec:backend}
\subsubsection{Data Overview.}
Starting from an initial cohort of 518 patients, we retained a curated cohort of 505 patients diagnosed with Stage II or Stage III cancer at an academic medical center. This research was conducted under the oversight of the University of California Institutional Review Board. The patient records were fully de-identified and analyzed as a secondary dataset by authorized data analysts; the research team did not directly interact with patients or access identifiable patient information. For the expert-review sessions described in \autoref{sec:evaluation}, we obtained informed consent from all participants, and the sessions were recorded with their consent.


The cohort spans three cancer types: breast (212 patients), non-small cell lung (150 patients), and colorectal (143 patients), where the colorectal cohort combines colon ($n = 100$) and rectum/rectosigmoid ($n = 43$) cases into a single group that is analyzed together. Cohort-level survival distributions are summarized in \autoref{tab:cohort}.

\begin{table}[h]
\centering
\caption{Patient cohort composition and survival status}
{\small
\label{tab:cohort}
\begin{tabular}{lrrr}
\toprule
Cohort & Total & Alive & Deceased \\
\midrule
Breast & 212 & 185 & 27 \\
Lung & 150 & 89 & 61 \\
Colorectal & 143 & 119 & 24 \\
\midrule
Total & 505 & 393 & 112 \\
\bottomrule
\end{tabular}
}
\vspace{-0.15in}
\end{table}

For each included patient, the dataset contains two categories of information: (1) basic clinical and demographic variables, including age, gender, treatments, comorbidities, and survival outcome, and (2) timestamped EHR access logs spanning a fixed 15-month analysis window, from three months before diagnosis to one year after diagnosis. A timestamped EHR access event involves a healthcare professional (HCP) accessing (reviewing or writing) a document, such as a note or a message, in the EHR system. Since HCPs typically record a patient's medical conditions in notes, while messages often lack sufficient context, we include only access events involving notes in our analysis. To support a consistent 15-month analysis window, we retain only patients whose EHR access logs cover the required period; patients whose records were insufficient for this window, including individuals who passed away too early after diagnosis, were excluded during cohort construction.

For each note, we retain attributes describing its intent category, content category, and inpatient status. For each HCP, we retain role-related attributes such as title, type, specialty, and resident status. These attributes are used both to characterize the collaboration networks described below and to derive interpretable summaries for the surrogate-based explanations.
\subsubsection{Care Team Network Construction. }
We represent each patient's EHR-mediated collaboration as a \emph{bipartite note-HCP graph}, where notes and HCPs serve as the two node types and edges represent reviewing or writing events recorded in the EHR. Accordingly, edges exist only between notes and HCPs, not between two HCPs or between two notes. This bipartite formulation reflects the fact that interactions among HCPs in the EHR are mediated through clinical notes rather than through direct provider-to-provider communication. These cumulative bipartite note-HCP graphs are the input representation for the temporal graph model introduced in \autoref{sec:model}.

To capture how collaboration evolves over time, we construct weekly cumulative snapshots. Using the patient's earliest recorded interaction within the 15-month analysis window as the temporal reference point, we partition the timeline into fixed-length one-week intervals. For each week $t$, the snapshot $G_t$ contains every node and interaction recorded from this reference point through the end of week $t$, producing a cumulative sequence $\{G_0, G_1, \dots, G_{T-1}\}$. This cumulative representation preserves when providers first enter the collaboration record while allowing the care team structure to grow over time.


This bipartite formulation has limitations that bound the interpretation of the resulting networks. EHR access logs capture only interactions mediated through the record; communication that occurs outside the EHR---such as verbal discussion, paging, or messaging not tied to note access---is not represented. Moreover, who appears in the record depends on local documentation and access practices, so we treat these graphs as a proxy for EHR-mediated collaboration rather than a complete record of care-team interaction (\autoref{sec:discussion}).

\subsection{Model Pipeline}
\label{sec:model}
\subsubsection{Temporal Prediction with EvolveGCN. }

To model how EHR-mediated collaboration evolves over time, we employ \textbf{EvolveGCN}~\cite{pareja2020evolvegcn} over the sequence of weekly cumulative bipartite note-HCP snapshots. EvolveGCN applies a graph convolution within each snapshot and uses a recurrent unit to evolve the convolution weights themselves from one snapshot to the next. Because the temporal signal is carried by the parameters rather than by persistent per-node embeddings, the model does not require every note or provider to be present in every week. This property matters here: as treatment progresses, new notes and new HCPs enter the record, so both the node set and the edge set grow across the sequence. Compared with static or temporally aggregated graph approaches, this formulation also preserves the ordering of the snapshots and the collaboration history accumulated up to each week.

EvolveGCN has two variants that differ in the role the convolution weights play inside the recurrent unit. In EvolveGCN-H, the weight matrix is the hidden state of a gated recurrent unit (GRU) whose input is the node embeddings of the current snapshot, so the weight update depends on node features. In EvolveGCN-O, the weight matrix is instead the input and output of a long short-term memory (LSTM), and node embeddings are not used in the recurrent update at all. The choice of model variant is dataset-dependent: the -H variant is preferable when node features are informative, and the -O variant when the temporal signal lies mainly in how the graph structure changes~\cite{pareja2020evolvegcn}. In our setting, a node's attributes are fixed categorical descriptors that do not change once that note or provider enters the record; what changes from week to week is which notes and providers are present and how they are connected. We therefore use EvolveGCN-O. For readability, we refer to this model simply as EvolveGCN in the remainder of the paper.

We train one EvolveGCN model per cancer cohort for binary survival classification, with the deceased outcome treated as the positive class. Unlike our prior full-timeline prediction setting~\cite{huang2026dynamics}, the temporal prediction setup here is organized at the patient-week level. For each week $t$, the model uses only the cumulative snapshot sequence $\{G_0,\dots,G_t\}$, so information from later weeks is not included in the week-$t$ prediction. The recorded patient outcome provides the supervision for each patient-week example. Repeating this process across weeks produces the sequence of model-estimated risks used throughout the visual analysis. Additional training details, including class-imbalance handling and optimization settings, are provided in the supplemental material (Appendix A).

\subsubsection{Surrogate-Based Feature Attribution.}
\label{3.3.2}
Although EvolveGCN produces a risk estimate for each week, its predictions are difficult to interpret directly. We therefore train a logistic regression surrogate using simpler, interpretable HCP and note attributes. The surrogate performs the same binary survival-prediction task as EvolveGCN but operates in an interpretable feature space. Each training example represents one patient at one week. For week $t$, the input summarizes the cumulative collaboration attributes observed through that week, while the patient's recorded survival outcome provides the training label. The surrogate produces a corresponding probability of the deceased outcome for each patient-week. Its learned coefficients are then used to identify which attributes are associated with higher or lower predicted risk.


Given the surrogate coefficient $\beta_j$ for feature $j$ and observed cumulative value $x_{j,t}$ at week $t$, the feature contribution is computed as $c_{j,t} = \beta_j x_{j,t}$. Positive contributions indicate features associated with higher model-estimated risk under the surrogate; negative contributions indicate features associated with lower model-estimated risk. For each patient-week, we retain the 80 underlying features with the largest $|c_{j,t}|$ for downstream visualization.

Because the surrogate performs the same survival-prediction task as EvolveGCN using a simpler, interpretable feature representation, it serves as an interpretable proxy for examining which collaboration attributes are associated with higher or lower predicted risk. Its coefficients should not be interpreted as a direct decomposition of EvolveGCN or as evidence of a causal clinical effect. The stability of the surrogate explanations across retraining is reported in the supplemental material (Appendix A).

The surrogate additionally supports an exploratory what-if analysis. When a user perturbs a selected feature, we use the learned surrogate coefficient to estimate the direction and magnitude of a corresponding change relative to the original EvolveGCN model-estimated risk. This calculation uses the surrogate coefficient to provide a directional projection anchored to the original EvolveGCN risk estimate.

\begin{equation}
\hat{p}_t = \sigma\!\left(\log\frac{p_t}{1 - p_t} - \beta_j \cdot \tfrac{\delta}{100} \cdot |x_{j,t}|\right)
\label{eq:perturbation}
\end{equation}
where $p_t$ is the original EvolveGCN risk at week $t$, $\delta$ is the user-specified perturbation percentage, and $\sigma$ is the sigmoid function. The resulting projected trajectory is intended only for exploratory reasoning about potential directional sensitivity in the learned model representation.

\subsection{Visual Analytics Design}
\label{sec:frontend}

\subsubsection{Design Rationale}


We follow a task-driven visual analytics design process~\cite{munzner2009nested}, abstracting domain needs into data and task descriptions that guide our visual encoding and interaction choices. At the domain level, clinicians and health-services researchers need to reason about how care-team activity evolves over time, how these changes align with model-estimated patient risk, and which care-team characteristics are emphasized by model explanations. As discussed in \autoref{sec:relatedworks}, prior work has largely kept temporal collaboration exploration, model-estimated risk, and feature-level explanation in separate workflows rather than integrating them on a shared patient-specific temporal basis.

The output of our modeling pipeline is high-dimensional and multimodal. For each patient, the system includes a temporal risk trajectory derived from EvolveGCN, a sequence of weekly cumulative bipartite note-HCP collaboration networks, per-week feature attributions with up to 80 underlying features, and weekly HCP activity snapshots identifying which care team members accessed the patient's notes during each interval. These outputs provide complementary perspectives on patient-specific care-team dynamics, but they remain difficult to interpret without a coordinated analytical workflow.

This work was developed in collaboration with clinical domain experts as part of an ongoing multi-team research effort; experts contributed throughout, from defining the domain terminology and provider-specialty grouping used across the system to shaping the analytical needs. Early in this process, clinical collaborators demonstrated the EHR interface used in their current practice (Epic Systems, a widely used EHR platform), in which information relevant to a patient's care team is spread across dense, list- and table-oriented screens that require extensive navigation and offer little support for tracking how collaboration and risk evolve together. We abstracted these needs through discussions with clinical and informatics collaborators and iterative refinement of the case-analysis workflow, yielding tasks that target these gaps: (T1) identify patients or patient pairs with notable combinations of care-team size, specialty breadth, recorded outcome, and model-estimated risk; (T2) inspect temporal risk trajectories and locate weeks with sustained high risk or sharp week-to-week changes; (T3) relate selected risk-transition weeks to care-team composition, feature-level surrogate explanations, and HCP co-access structure; and (T4) compare patients side by side to generate hypotheses about how different collaboration patterns may be associated with different model-estimated risk trajectories.

These tasks motivated a coordinated multi-view design rather than a single unified visualization. We considered linear timelines for risk trajectories, matrix-based representations for HCP co-access, and aggregate statistical summaries of care-team structure. A linear timeline can support more precise value comparison, and prior work has shown that linear charts can outperform radial charts for many time-based reading tasks~\cite{8807238}. However, our use of the radial glyph is not intended for exact value reading; instead, it provides a compact overview of patient-level trajectory shape, supports rapid identification of temporal transitions, and enables side-by-side comparison while preserving space for linked interpretation views. Exact values and week-specific inspection are supported through linked interaction and downstream panels.

For the collaboration structure, an adjacency matrix could scale better for dense networks~\cite{ghoniem2004comparison} and aggregate summaries could provide compact statistics, but both make it harder for clinical collaborators to inspect local provider neighborhoods and specialty-integrated co-access patterns. We therefore use a patient-centered ego-network node-link representation~\cite{munzner2014vad} for weekly and cumulative co-access inspection. This representation supports the central analytical task of understanding which HCPs appear connected through shared note activity, how specialty groups are positioned relative to one another, and how the local structure changes across selected weeks. The resulting design prioritizes coordinated hypothesis generation: the radial glyph identifies when to inspect, the attribution and distribution views explain which care-team attributes are emphasized, and the ego-network view shows how those providers are structurally organized. We emphasize that the visualization contribution lies not in a new standalone chart type but in this task-driven coordination of temporal graph modeling, patient-level risk trajectories, surrogate feature attribution, and HCP co-access structure on a shared weekly collaboration representation, enabling analysts to move coherently between when risk changes, which care-team attributes are implicated, and how the corresponding care-team structure is organized. Guided by these tasks, the interface is organized around six coordinated views, each targeting a distinct analytical task within the workflow described next.

\subsubsection{Workflow Overview}

While \autoref{fig:flowchart} summarizes the system components and their outputs, the analyst-facing workflow is organized into four steps. Each step narrows the analytical focus and establishes the context for the next.

\textbf{Step 1: Cohort Overview and Patient Selection.}
Analysts begin at the cohort level, filtering on cancer cohort and survival status to locate patients or patient pairs whose combination of care-team size, specialty breadth, recorded outcome, and model-estimated risk warrants closer examination (T1). The outliers and comparison candidates identified here become the cases carried through the remaining steps.

\textbf{Step 2: Temporal Inspection and Week Selection.}
For a selected patient, analysts locate the weeks that warrant closer inspection, whether periods of sustained high risk or points of sharp week-to-week change (T2). Because absolute risk and week-to-week change surface different weeks, both readings are available, and the week selected here establishes the temporal focus for the remaining steps.

\textbf{Step 3: Care-Team Interpretation.}
Analysts then relate the selected week to the care team behind it, examining which collaboration attributes the surrogate explanation emphasizes, which specialties are active, and how providers are connected through shared note access (T3). Moving across weeks supports comparative inspection of how these change over the treatment timeline, and comparing two patients at the same relative time point supports hypotheses about how different collaboration patterns may be associated with different model-estimated risk trajectories (T4).

\textbf{Step 4: Exploratory What-If Analysis.}
Finally, analysts may examine whether an emphasized attribute plausibly relates to the modeled risk, by observing how the surrogate projection responds to a hypothetical reduction in that attribute. The result is a hypothesis for further investigation rather than a clinical claim.
\subsubsection{View Design}

\begin{figure*}[t]
    \centering
    \includegraphics[width=\textwidth]{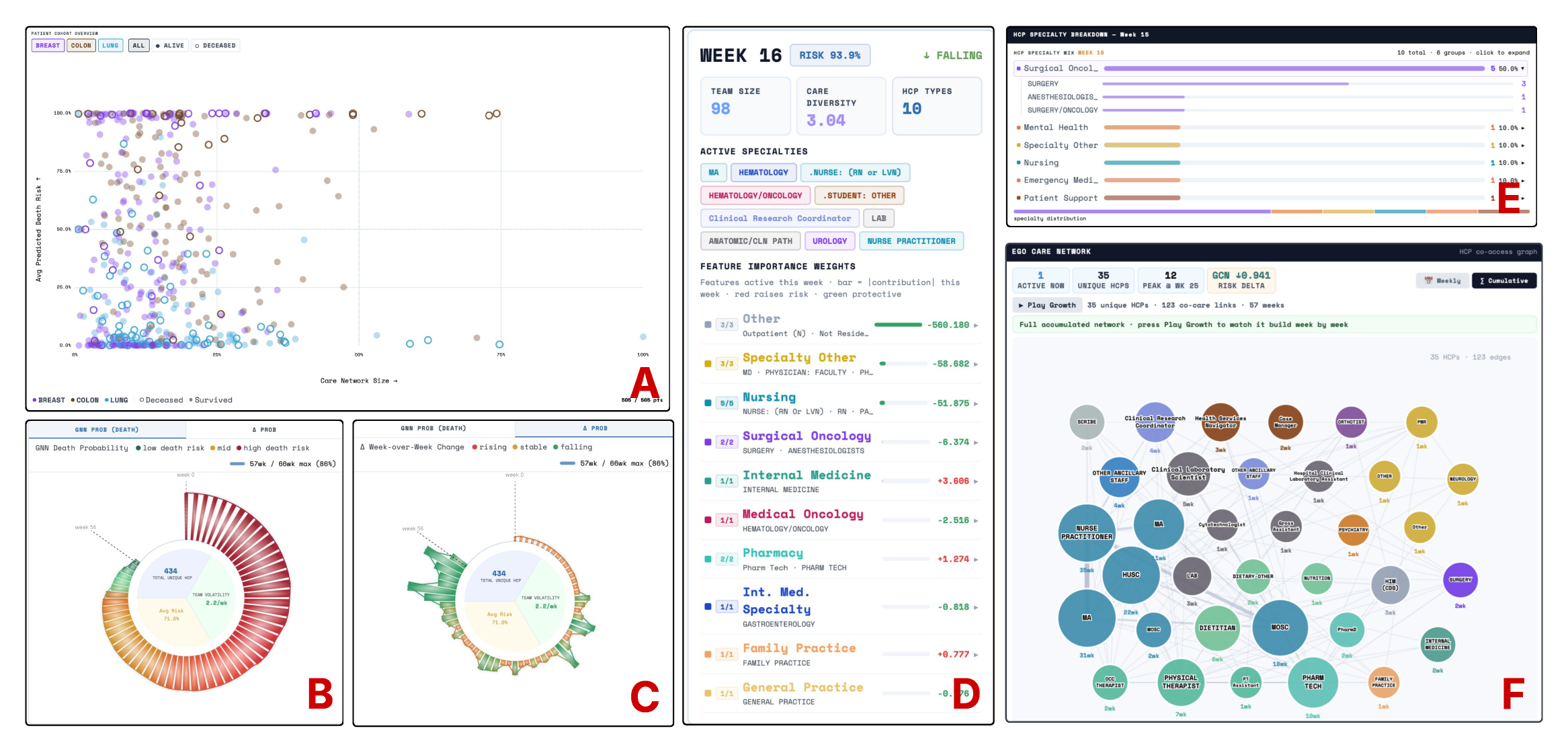}

    \caption{Composed overview of the coordinated views in the visual analytics system, including cohort-level selection (A), temporal risk inspection (B and C), feature attribution (D), specialty composition analysis (E), and collaboration-structure exploration (F). 
    }
    \label{fig:system_overview}
\end{figure*}

\autoref{fig:system_overview} presents a composed overview of the coordinated views used for cohort-level selection, temporal risk inspection, feature attribution, specialty composition analysis, and collaboration-structure exploration. We describe each view and its encoding below.

\textbf{Cohort Scatterplot (Step 1, T1; panel A of \autoref{fig:system_overview}).}
This view serves as the entry point for cohort-level exploration. Each point represents one patient. The x-axis encodes care-network size, defined as the total number of unique HCPs who accessed the patient's notes and normalized as a percentage of the cohort maximum. The y-axis encodes the patient's average model-estimated risk across all weekly snapshots. Point color indicates cancer type (purple for breast, brown for colorectal, and blue for lung), while fill style indicates survival outcome (filled circles for surviving patients and hollow circles for deceased patients). Together these encodings let care-team size, model-estimated risk, cancer type, and recorded outcome be read at a glance.


In addition to the default cohort projection, the overview scatterplot serves as a flexible cohort-level entry point whose axes can be reassigned to other patient-level summaries depending on the analytical question. For example, the x-axis can encode cumulative specialty breadth, defined as the number of distinct HCP specialties involved across the care trajectory, while the y-axis can encode peak risk escalation, defined as the largest positive week-to-week change in model-estimated risk. This configuration can surface patients with sharp changes in model-estimated risk across different levels of care-team specialty breadth, supporting case selection for subsequent longitudinal analysis.

\textbf{Radial Glyph (Step 2, T2; panels B and C of \autoref{fig:system_overview}).}
This view encodes the selected patient’s temporal model-estimated risk trajectory as a radial bar chart centered on a summary disc, where each bar represents one weekly snapshot. A thin overlaid curve traces the outer bar profile to provide a continuous visual guide to the overall trajectory shape across weeks. The glyph supports two toggleable modes to emphasize different aspects of temporal change. In \textsc{Probability} mode, bar height reflects the model-estimated risk for that week, and color follows a low-to-high risk gradient from green to red to provide an immediate sense of absolute risk over time. In \textsc{Delta} mode, bar height reflects the magnitude of week-to-week change in risk, while color indicates direction, distinguishing increasing, relatively stable, and decreasing risk. Rather than always forming a full circle, the glyph spans an angular range proportional to the patient’s observed timeline, allowing viewers to distinguish patients with longer versus shorter observed treatment timelines. The center disc summarizes patient-level statistics across the full trajectory. Selecting any bar sets the temporal focus and synchronizes the linked interpretation views to the same week.


\textbf{Surrogate Attribution Panel (Step 3, T3; panel D of \autoref{fig:system_overview}).}
This view displays the surrogate linear model’s per-feature contributions at the selected week, where each feature reflects cumulative collaboration attributes. Each displayed feature combines two underlying components: a \texttt{::freq} count capturing how often that activity has occurred and a \texttt{::present} indicator capturing whether it has appeared at all. Features are shown as horizontal bars sorted by absolute contribution magnitude, with bar length indicating contribution strength and color indicating direction, where red denotes a risk-increasing contribution and green denotes a risk-decreasing contribution. To support interpretation, the panel also shows the cumulative feature information and corresponding surrogate coefficients beneath each active feature bar, helping analysts distinguish between features that stand out because they occur frequently and those that reflect stronger model sensitivity.

\textbf{HCP Distribution Chart (Step 3, T3; panel E of \autoref{fig:system_overview}).}
This view presents a grouped horizontal bar chart showing the distribution of HCPs across a two-level specialty taxonomy, including 22 top-level groups and approximately 150 fine-grained specialties. The top-level categories were defined in collaboration with clinical domain experts to organize the full set of EHR-recorded provider specialties into clinically meaningful groups such as Medical Oncology, Surgical Oncology, Nursing, Pharmacy, and Internal Medicine Specialties. The chart can display specialty composition for the selected week or in aggregate across all weeks, allowing analysts to examine how the functional makeup of the care team changes over time and differs across patients or treatment stages. Hovering over a top-level group reveals the corresponding fine-grained specialty breakdown for more detailed inspection.

\textbf{Ego Network (Step 3, T3; panel F of \autoref{fig:system_overview}).}
This view visualizes an HCP-HCP co-access network derived from the patient’s bipartite note-HCP graph, where two providers are connected when they both accessed the same patient note, making the note the implicit intermediary of the relationship. Its purpose is to help analysts interpret the structural organization of the care team beyond feature rankings and specialty counts by showing which providers appear connected through shared note activity and how those relationships evolve over time. The view supports two complementary modes. In \textsc{Weekly} mode, it shows the co-access structure for a selected week, allowing analysts to inspect whether provider interactions appear sparse, fragmented, or more tightly connected at that point in the care timeline. A scrubber spans the full observation window and enables week-by-week navigation, with blue markers indicating weeks in which HCP-HCP co-access connections were observed and gray markers indicating weeks with provider activity but no co-access ties. For weeks in which providers accessed notes without forming shared-note connections, the view falls back to a simplified layout that places active HCPs around a central patient node, preserving evidence of provider activity even when no co-access network is present. In \textsc{Cumulative} mode, the view aggregates all weekly snapshots into a single network to summarize longer-term collaboration patterns; node size encodes how many weeks a provider was active, edge thickness encodes how often a provider pair co-accessed notes across weeks, and node color follows the specialty grouping used in the HCP distribution chart, including split-color arcs for providers associated with multiple groups. Key interactions support both temporal and structural analysis: the scrubber and play control animate changes over time, hovering reveals specialty details, and selecting a node highlights its immediate neighborhood while dimming unrelated parts of the graph.

\begin{figure}[t]
    \centering
    \includegraphics[width=\columnwidth]{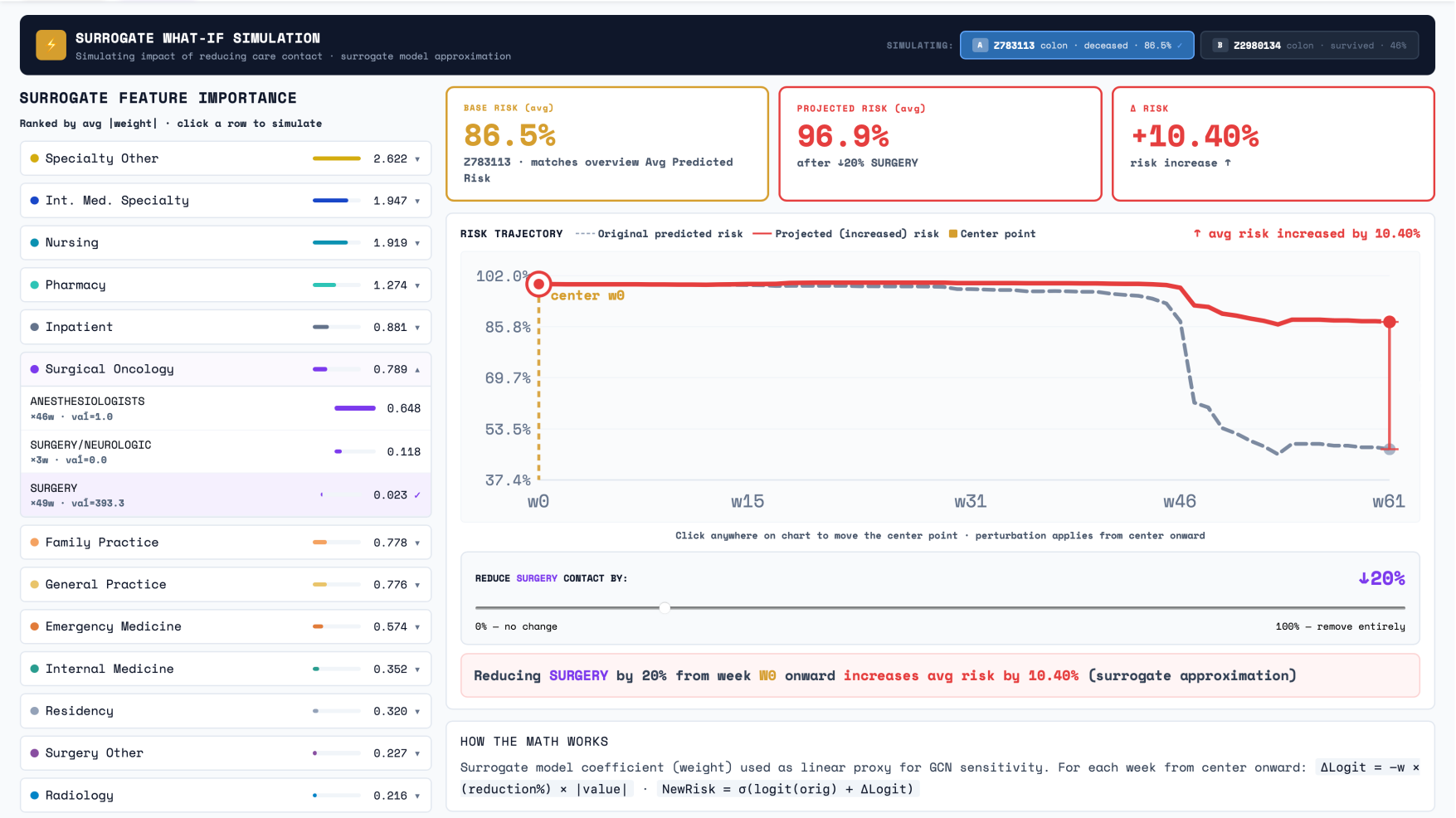}
    \caption{Surrogate-based what-if analysis of surgery-related involvement. Reducing surgery-related activity by 20\% from week~0 shifts the projected average risk upward under the surrogate approximation.}
    \label{fig:whatif_case}
\end{figure}

\textbf{What-If Panel (Step 4; \autoref{fig:whatif_case}).}
This view supports exploratory what-if analysis through surrogate-based feature perturbation. Analysts select a feature from the Surrogate Attribution Panel and adjust a perturbation slider to simulate a 0--100\% reduction in that feature. The system then computes a surrogate-projected risk trajectory using \autoref{eq:perturbation} and overlays it on the original model-estimated trajectory, allowing analysts to compare possible directional changes under the selected perturbation. Because the perturbed trajectory comes from the linear surrogate rather than from re-running EvolveGCN, it serves as a hypothesis-generation tool for exploring model sensitivity in an interpretable feature space, not a clinical prediction or recommendation.

%% file: 4_case_study.tex
\section{Case Studies}
\label{sec:casestudy}

We illustrate the analytical workflow through two case studies drawn from the two colorectal cancer cohorts. These cases show how an analyst can use coordinated views to investigate specific questions about how model-estimated risk relates to patient-specific collaboration patterns over time. In each case, the interpretation is developed by linking evidence across different views. The resulting observations are intended to support insight generation and hypothesis development, and should therefore be understood as exploratory patterns surfaced through the visual analytics system rather than as prescriptive clinical conclusions.

These case studies were developed in conjunction with the expert demonstration sessions described in \autoref{sec:evaluation}: we presented the system to domain experts, whose feedback confirmed which interpretations were clinically plausible and directly shaped the clinically constrained comparison strategy in \autoref{sec:case2}.
\subsection{Exploratory Comparison: Divergent Risk at Similar Care-Network Size}
\label{sec:case1}

We begin with two colorectal cancer patients selected from the cohort scatterplot, where the x-axis encodes care-network size and the y-axis encodes average model-estimated risk across the full timeline. They were selected as an exploratory comparison pair because they have similar care-network size but substantially different average model-estimated risk (86.5\% versus 46.0\%), as shown in the cohort overview, allowing us to examine how patients with comparable network scale can nevertheless differ in temporal risk patterns and collaboration structure. The recorded outcomes also differ: the labeled deceased patient is predicted to be at a higher risk, whereas the labeled surviving patient is predicted to have a lower risk.

\begin{figure}[t]
    \centering
    \includegraphics[width=\columnwidth]{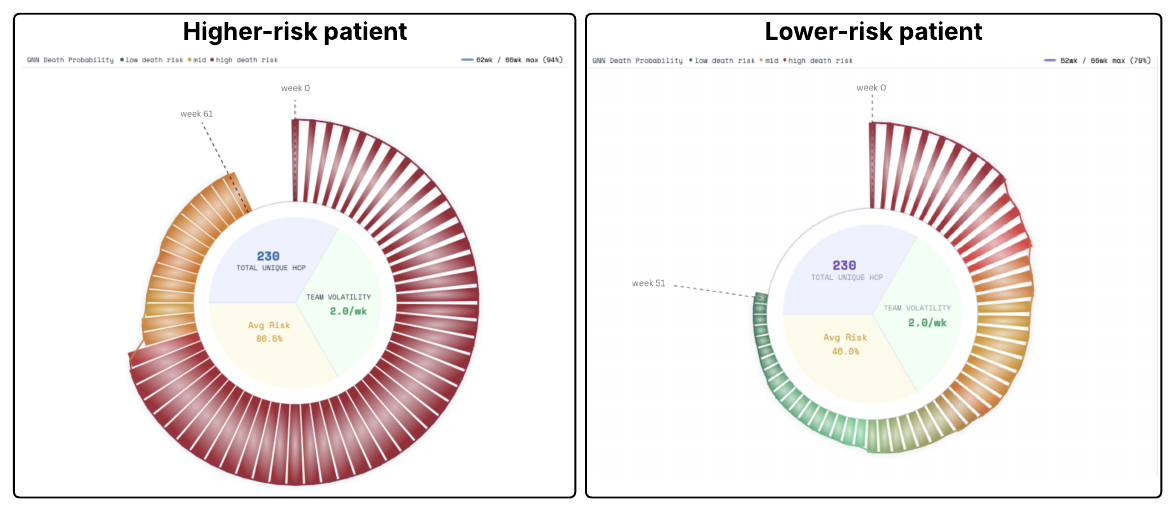}
    \caption{GNN Probability-mode radial comparison for the selected patient pair
    in \autoref{sec:case1}.}
    \label{fig:case1_gnn}
    \vspace{-0.2cm}
\end{figure}

\autoref{fig:case1_gnn} shows the radial glyphs of the chosen patient pair under \textsc{Probability} mode. The higher-risk patient's trajectory remains largely red throughout the timeline, with only a modest shift toward orange near the end, suggesting persistently elevated model-estimated risk. In contrast, the lower-risk patient's trajectory gradually transitions from red toward green, suggesting a steady improvement over the course of treatment. The center summary shows that both patients have the same total number of unique HCPs (230) and the same average team volatility (2.0/week; defined in the supplemental material, Appendix B), suggesting that the difference in model-estimated risk is not explained solely by aggregated statistics of team composition.

\begin{figure}[b]
\centering
\includegraphics[width=\columnwidth]{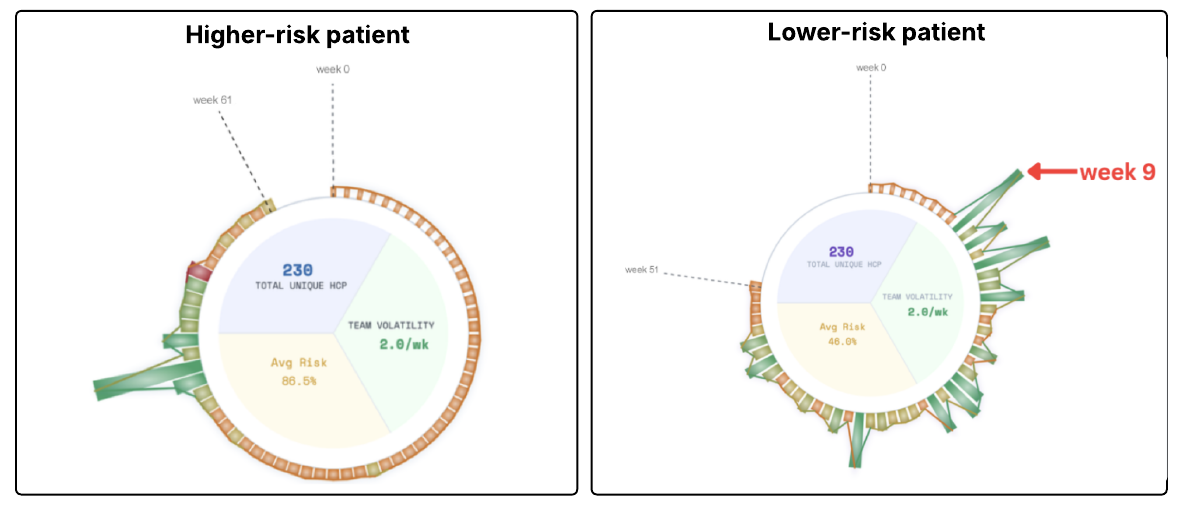}
\caption{ Delta-mode radial comparison for the \autoref{sec:case1} patient pair. The higher-risk patient (left) shows persistent orange spikes with little green activity, while the lower-risk patient (right) shows stronger downward risk shifts, particularly around week~9.}
\label{fig:case1_radial}
\vspace{-0.10in}
\end{figure}

\autoref{fig:case1_radial} shows the pair in \textsc{Delta} mode. The higher-risk patient is dominated by small orange spikes with relatively little green activity, whereas the lower-risk patient shows multiple green spikes throughout the entire treatment course, indicating gradual week-to-week declines in estimated risk over time. We focus on week~9 to investigate what contributed to this earliest pronounced drop in the lower-risk patient’s estimated-risk trajectory.


In the week~9 Surrogate Attribution Panel, the lower-risk patient exhibits a broader set of prominent risk-lowering features, including \textit{Nursing}, \textit{Internal Medicine}, \textit{Internal Medicine Specialty}, and \textit{Radiology}. In contrast, the higher-risk patient shows a narrower feature profile and includes a positive contribution from \textit{Family Practice}. Complementing these observations, the summary of \emph{active} HCPs---defined as those with at least one recorded interaction in week~9---indicates that the higher-risk patient has a smaller and less diverse care team (22 providers across 2 HCP types), whereas the lower-risk patient involves 36 providers spanning 11 HCP types.
Together, these differences suggest that the lower-risk patient’s improvement in the week~9 risk estimate is associated with a more diverse and actively engaged set of HCPs aligned with lower-risk collaboration patterns. Conversely, the higher-risk patient’s collaboration at week~9 involves fewer and less diverse HCP roles linked to favorable outcomes, which may help explain the absence of a corresponding decrease in predicted risk.

\begin{figure*}[t]
\centering
\includegraphics[width=0.8 \textwidth]{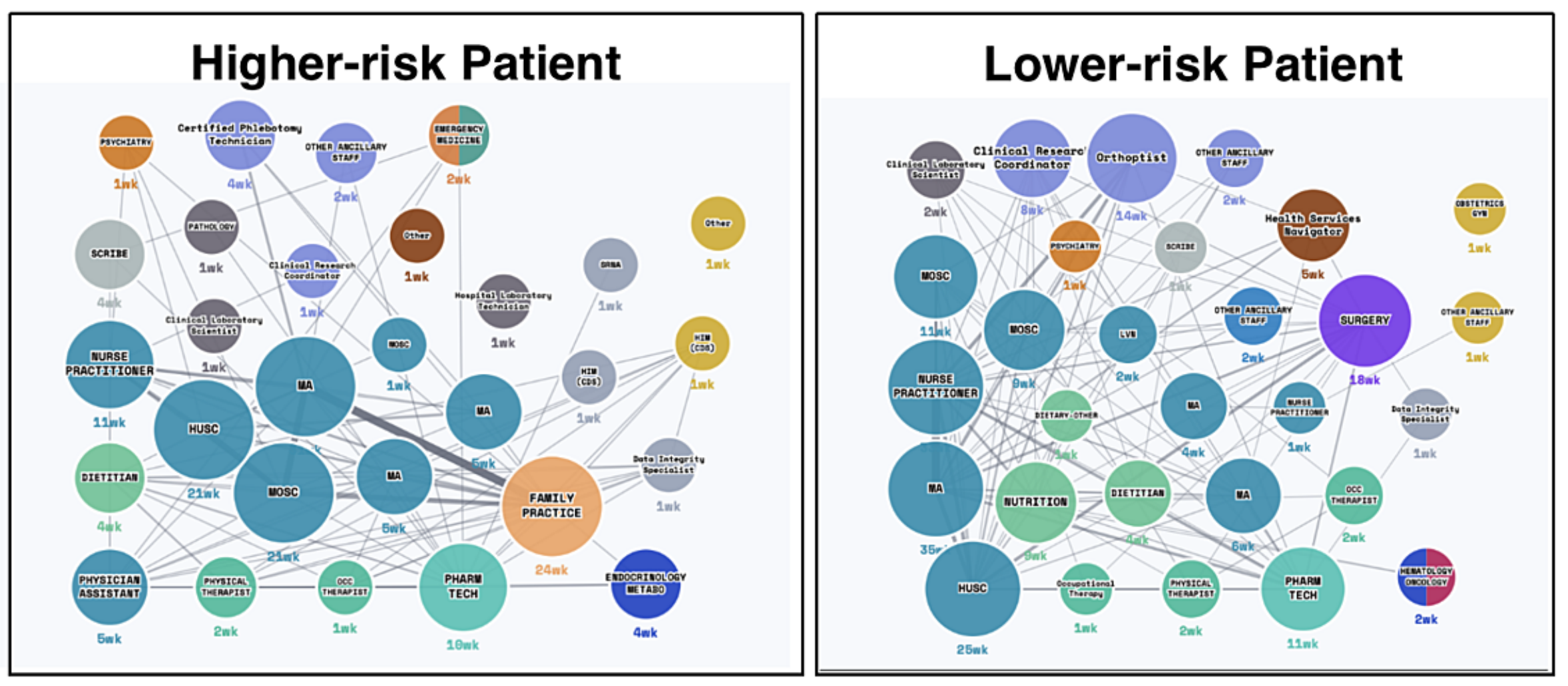}
\caption{Cumulative ego-network comparison of the two \autoref{sec:case1} colorectal patients. Each node is a healthcare professional (HCP), labeled by role or specialty; node size and the accompanying week count encode how many weeks that provider was active in the patient's record, and node color denotes the specialty group defined in the HCP distribution taxonomy (split-color nodes belong to multiple groups). An edge connects two providers who co-accessed the same clinical note, and edge thickness encodes how often that pair co-accessed notes across the treatment timeline---thicker edges mark provider pairs with more frequent shared-note co-access, and thus more repeated EHR-mediated interaction. The lower-risk patient (right) forms a more interconnected and more specialty-diverse network, with surgical and medical oncology tightly integrated, whereas the higher-risk patient (left) is sparser and less integrated around surgery-related specialties.}
\label{fig:case1_ego}
\vspace{-2pt}
\end{figure*}

The cumulative ego network further reveals how actively participating HCPs are connected. As shown in \autoref{fig:case1_ego}, the key difference lies in the structure of these connections: the higher-risk patient has 83 co-access edges and a visually sparser network, whereas the lower-risk patient has 99 co-access edges and a more interconnected structure. Nursing remains the dominant specialty group for both patients; however, surgical oncology and medical oncology are more tightly integrated in the lower-risk patient’s network. In contrast, surgery-related specialties appear less integrated in the higher-risk patient’s collaboration structure. Together, these patterns suggest that the contrast between the two patients is not only in the presence of specific roles, but in how oncology- and surgery-related providers are positioned within the broader cumulative co-access network over time.

With surgery-related features surfaced by the surrogate as risk-lowering under the model approximation, we conduct a what-if analysis by applying a 20\% reduction in surgery-related activity starting from week~0 as a perturbation. As shown in \autoref{fig:whatif_case}, the average surrogate-projected risk for the deceased patient increases by approximately 10.4\%, rising from 86.5\% to 96.9\%. Applying the same perturbation to the surviving patient also leads to an increase in projected risk, from 46.0\% to 50.6\% (+4.60\%). Despite differences in outcome and overall risk trajectories, both patients exhibit a consistent pattern: under this surrogate perturbation, reducing surgery-related activity increases the projected risk for both patients. Taken together, this case study links structural observations with what-if analysis to support hypothesis generation for further investigation.

\subsection{Expert-Guided Clinically Constrained Comparison}
\label{sec:case2}

The first case study demonstrates how the system supports detailed interpretation of divergent risk trajectories through coordinated temporal, attribution, and network views. However, selecting patients primarily by similar care-network size leaves open an important clinical concern: observed differences may reflect underlying differences in disease context rather than care-team organization. In the expert-feedback sessions, clinicians emphasized that patient comparisons would be more informative when patients share major clinical characteristics, such as cancer type and stage. We therefore use this second case study to demonstrate a more clinically constrained comparison workflow, in which the system is used to examine whether patients with similar clinical context and aggregate EHR activity still differ in the temporal organization of provider activity.

Motivated by this feedback, and to examine whether a similar structural contrast appears beyond the primary institution, we applied the same workflow to a colorectal cohort independently collected at an additional site; details of this cohort and its data collection are provided in the supplemental material (Appendix C, Table 1). To make the comparison more clinically constrained, we selected two patients at the same cancer stage (Stage 2B) with comparable care-team scale and documentation volume, but different recorded outcomes. Patient A, who was deceased, had a substantially higher average model-estimated risk of 96.6\%, with 272 HCPs and approximately 22K notes. Patient B, who survived, had a much lower average model-estimated risk of 6.1\%, with 269 HCPs and approximately 21K notes. Despite this difference in risk, the two patients appear similar at the aggregate level—comparable provider counts and documentation volume—allowing us to examine whether linked temporal and network views reveal differences in care-team organization not apparent from aggregate counts alone.

Using the radial risk view, we selected a week with a pronounced change in the lower-risk patient’s trajectory and inspected the linked weekly care-team views. During this week, Patient B showed 8 active HCPs and 9 co-access edges spanning 7 HCP types, whereas Patient A showed 4 active HCPs and only 1 co-access edge. The surrogate attribution view also highlighted multiple specialty-related features associated with lower model-estimated risk for Patient B, while the weekly summary indicated higher note activity and slightly higher care diversity. These linked views show that the difference between the patients was not simply the number of providers or notes, but how provider activity was temporally organized and connected around a key point in the risk trajectory.

The cumulative network view provided a broader view of this contrast. Patient A's cumulative network spanned 32 distinct specialties across 40 co-access edges over 58 weeks, whereas Patient B's spanned 28 specialties across 61 co-access edges over 51 weeks. Thus, despite a narrower specialty range and a shorter observed timeline, Patient B showed more frequent co-access relationships among providers. This case illustrates the value of the network view for distinguishing care-team structure from care-team scale: two patients may appear similar in aggregate EHR activity, yet differ substantially in the density and specialty composition of their temporal co-access patterns.

This case study contributes a distinct use scenario from the first case. Rather than only surfacing a potentially meaningful pattern in a selected patient pair, it shows how domain-expert feedback can shape the comparison strategy itself. Reproducing the \autoref{sec:case1}-style structural contrast on an independently collected cohort also provides preliminary evidence that the observation is not an artifact of a single site. The system supports a workflow in which clinicians or analysts first constrain the comparison to clinically similar patients, then use linked temporal and network views to identify more specific hypotheses about coordination patterns. These observations remain exploratory and should not be interpreted as evidence that denser connectivity directly improves survival. Expert feedback emphasized that increased EHR-mediated activity may also reflect complications, hospitalization, or greater clinical complexity. Therefore, this case motivates follow-up cohort-level analyses that account for comorbidities, treatment pathway, hospitalization history, and timing of major clinical events.

%% file: 5_expert_feedback.tex
\section{Evaluation}
\label{sec:evaluation}

We evaluated the system through two complementary forms of evidence. First, we conducted a qualitative expert review to assess whether the coordinated views support clinically plausible interpretations of temporal care-team dynamics and to identify constraints on how such analyses should be framed. Second, we used prior clinical literature to contextualize the care-team patterns surfaced in the case studies and explain why they are clinically relevant to investigate, while avoiding causal interpretation. Together, these activities assess the system's interpretive usefulness and provide clinical context for hypotheses surfaced through exploratory analysis, rather than treating them as definitive clinical conclusions.

\subsection{Expert Review Protocol}

The expert review was conducted as two live demonstration and discussion sessions as part of an ongoing multi-team, multi-institutional cancer-care research collaboration. These sessions brought together domain experts actively involved in the project, reflecting a coordinated effort across clinical, informatics, and health systems research teams. The first session included 13 participants and the second 11, with some experts attending both. Participants included clinicians (e.g., physicians in internal medicine, oncology, surgery, and hospital medicine), medical informatics researchers, health services researchers, and experts in organizational psychology and healthcare administration, along with members of the visualization and machine learning team. The participants had expertise in cancer care delivery, EHR-mediated communication, multi-team systems, clinical informatics, and statistical analysis of care-team networks.

Each session began with a brief overview of the visual analytics system. We demonstrated how users can move from the cohort overview to patient-level risk trajectories, select a time period of interest, and inspect linked views of surrogate attribution, specialty composition, and HCP co-access structure. The demonstration used a patient-comparison scenario designed to support assessment of whether the visualized patterns were clinically interpretable, what analysis tasks the tool could realistically support, and what additional clinical context would be needed before drawing stronger conclusions.

Following each live demonstration, we held a discussion with the participating experts to gather their feedback on the system’s potential use in clinical and research settings, the interpretability of the observed care-team differences, and areas where additional context or functionality would be needed.

\subsection{Expert Feedback Themes}

The two sessions were conducted over Zoom and recorded with participant consent. We transcribed the recordings and organized the discussion into recurring themes through iterative review by the research team, and each theme is summarized using generalized expert labels rather than named attributions to preserve anonymity.

The feedback in this section draws from both expert-review sessions. Feedback from the first session primarily informed the interpretation of \autoref{sec:case1}, while feedback from the second session informed the refined comparison strategy used in \autoref{sec:case2}.

\textbf{Clinically constrained comparison.} Following the first live demonstration, Clinician Expert A emphasized that patient comparisons would be more convincing if the selected patients were matched on major clinical characteristics before interpreting differences in care-team structure. In response to the colorectal comparison in \autoref{sec:case1}, the expert asked whether the patients were matched by cancer stage, cancer type, time from diagnosis to first therapy, and other clinical factors. The expert further suggested that, when presenting such comparisons to clinicians, it is important to show that patients are comparable in terms of disease severity and treatment context; otherwise, observed differences in model-estimated risk may be driven by underlying clinical differences rather than care-team dynamics. 
Suggested matching variables included age, gender, cancer stage, treatment pathway, time to first treatment, and type of first therapy. This feedback directly motivated the refined comparison strategy used in the second session and in \autoref{sec:case2}: rather than selecting patients only by similar aggregate care-network size, the follow-up demonstration selected patients with the same cancer stage and similar overall EHR activity, to improve the clinical comparability of the selected patients. This refinement showed how domain-expert guidance can inform the selection of clinically comparable cases before temporal and structural drill-down.


Experts also asked to see patient comorbidities when interpreting selected cases. We therefore treat comorbidities as additional clinical context when constructing and interpreting patient comparisons. Prior analysis on this cohort further found that care-team collaboration carries predictive signal beyond comorbidity and disease severity, with collaboration-based models outperforming comorbidity-only models, and with clinical factors such as stage only weakly associated with survival \cite{lu2025associatinghealthcareteamworkpatient}, supporting the interpretation of care-team structure as complementary to, not a proxy for, disease severity.

\textbf{Clarifying the purpose of the visualization.} Clinician Expert B noted that the interpretation of the system depends on whether the goal is prediction or care-process exploration. If the goal is purely predictive, then survival and other outcomes may be strongly influenced by disease biology, treatment pathway, and patient factors outside the care-team network. However, if the goal is to understand how EHR-mediated care-team activity and organization relate to care delivery, then the visualization can help identify patterns that warrant further investigation. This distinction helped clarify the intended use of the system. Rather than positioning the interface as a tool for explaining survival outcomes directly, we frame it as a visual analytics system for exploring how model-estimated risk co-evolves with EHR-mediated care-team activity and for generating hypotheses about care coordination. Other clinical participants agreed that the system's usefulness depends on this prediction-versus-exploration distinction.

\textbf{Value of coordinated temporal drill-down.} Most participants across both sessions emphasized that the interface's value lies in its linked workflow across views rather than any single visualization, because the coordination lets them relate changes in model-estimated risk to changes in specialty composition and HCP co-access activity at the same time point. In practice, users can start from a patient-level risk trajectory, select a week with a notable change, and examine specialties, summary statistics, attributions, and co-access structure for that week.

Experts noted that synchronized views reduce cognitive load by preserving temporal context and reducing the need for manual cross-referencing. The ability to move between overview and detail supports hypothesis refinement: users can identify patterns in trajectories, examine them through composition and attribution signals, and examine network structure to assess coordination. This iterative process enables more rigorous questioning, such as distinguishing transient activity from sustained coordination.

Overall, the two sessions support framing the system as a domain-expert-guided hypothesis-generation tool: experts identify candidate coordination patterns through the interface, then test them in matched cohort analyses incorporating clinical variables and downstream care-delivery outcomes.

\subsection{Connecting Expert Feedback to Colorectal Care-Team Literature}


Because both case studies focus on colorectal cancer, we use prior literature on colorectal care and EHR-mediated collaboration to contextualize the observed patterns around clinical comparability, multidisciplinary coordination, treatment timing, and the limits of access-log data.

The experts' emphasis on clinically constrained comparison aligns with colorectal cancer care, where prognosis and treatment depend strongly on disease stage, treatment pathway, and timing of major interventions. This supports constraining patient comparisons by major clinical characteristics such as cancer stage, as done in \autoref{sec:case2}. Prior guidelines and reviews also identify surgery as a central component of colon cancer treatment, particularly for earlier-stage disease \cite{NCI,costaschavarri2019treatment,chakrabarti2020earlycolon}, providing clinical context for examining the surgery-related patterns surfaced in our analysis.

The experts' caution about communication patterns is also consistent with colorectal care literature. Multidisciplinary colorectal cancer care has been associated with improved survival, and treatment timing, including delays to surgery or postoperative chemotherapy, has been linked to patient outcomes \cite{peng2021mdt,whittaker2021delay,gao2018timing,yang2023optimal}. These findings make surgery-centered and treatment-phase collaboration patterns clinically plausible to inspect. However, they do not imply that denser EHR co-access networks directly improve outcomes. As experts noted, increased interaction with surgical teams, hospitalists, or emergency medicine providers may also reflect complications, hospitalization, or greater clinical complexity.

Finally, this feedback reinforces prior work using EHR access logs to study care-team collaboration. Access patterns have been used to construct patient-centered and virtual care-team networks and to examine information-sharing among providers \cite{yao2018ehr,Zhu2019-tu,zhu2022ehr}; our system extends this by linking access-log-derived structures with temporal model-estimated risks and surrogate explanations. As both the literature and experts note, access logs are digital traces of care activity, not direct evidence of communication quality, clinical intent, or causal effect—reinforcing the system's role as a tool for generating coordination hypotheses to be tested in matched colorectal cohorts with richer clinical variables and downstream care-delivery outcomes.

%% file: 6_discussion.tex
\section{Discussion and Limitations}
\label{sec:discussion}
Our work links temporal risk estimates, surrogate feature explanations, specialty composition, and HCP co-access structure to support comparative and longitudinal reasoning about how care-team activity changes alongside model-estimated risk. The case studies and expert feedback show that this workflow can help analysts move from aggregate patient comparison to temporally specific questions about care-team organization. However, the patterns surfaced by the system should be interpreted as exploratory signals for hypothesis generation rather than as clinical conclusions or evidence of causal effect.

A central limitation of our current approach is that the collaboration networks are derived from EHR access logs. These logs do not capture communication outside the record—face-to-face discussion, phone calls, secure messages, or informal coordination—nor do they encode clinical intent, decision-making context, or communication quality. Documentation practices, search behavior, AI-assisted documentation, and institutional workflow differences further affect which providers appear and how often. We therefore interpret the observed co-access networks as proxies for EHR-mediated collaboration, not complete representations of care-team interaction.

The case studies illustrate the analytical workflow rather than establish controlled clinical comparisons. The revised colorectal cases use clinically constrained comparisons, but stronger validations would require larger matched-cohort analyses incorporating richer clinical variables—comorbidities, treatment events, and laboratory/radiology activities—to determine whether the patterns hold after accounting for patient complexity.

The surrogate attribution and what-if views should also be interpreted cautiously: they provide interpretable feature-level associations and exploratory projections, but do not represent clinical interventions or causal effects. A simulated feature reduction probes model behavior in an interpretable feature space; it is not evidence that changing that provider's involvement would change patient outcome. Future work should examine whether similar signals remain stable under alternative temporal graph models, explanation methods, and feature definitions.

Finally, future validation should move beyond survival alone: endpoints more directly tied to care coordination—such as emergency-department use, readmission, and treatment delays—would better show whether the surfaced patterns reflect clinically meaningful differences in care delivery.

%% file: 7_conclusion.tex
\section{Conclusion}
\label{sec:conclusion}

We presented a visual analytics framework for investigating cancer care as a time-evolving collaboration process, aligning temporal graph-based risk modeling with the same evolving collaboration networks used in the visual analysis and coordinating risk trajectories, specialty composition, feature-level explanations, and HCP co-access structure within a coordinated workflow. Through case studies and expert feedback, we illustrated how this coordination helps analysts surface and compare patterns of care-team organization that are difficult to observe from static or aggregated summaries alone. Rather than supporting definitive clinical conclusions, the framework is intended for insight generation and hypothesis development, offering a foundation for human–AI analysis of complex, evolving care processes in oncology and related team-based healthcare settings.